\documentclass[letterpaper, 10 pt, conference]{ieeeconf}  

\usepackage{amsmath} 
\usepackage{amsfonts}
\usepackage{graphicx}  
\usepackage{booktabs}
\usepackage{multirow}
\usepackage{array}         

\IEEEoverridecommandlockouts                              

\title{\LARGE \bf
Learning Geometry-Aware Spatio-Temporal Representations for Depth-Enhanced Visual Relocalization
}

\author{Deben Lu$^{1\dagger}$, Wendi Yang$^{1\dagger}$, and Yunxiao Shan$^{1*}$
\thanks{$\dagger$These authors contributed equally to this work.}%
\thanks{* is the Corresponding author.}%
\thanks{$^{1}$Deben Lu, Wendi Yang, and Yunxiao Shan are with the School of Artificial Intelligence, Sun Yat-sen University, Guangzhou, China.{\tt\small \{ludb,yangwd\}@mail2.sysu.edu.cn, shanyx@mail.sysu.edu.cn}}%
\thanks{This work was not supported by any organization.}%
}

\begin{document}

\maketitle
\thispagestyle{empty}
\pagestyle{empty}

\begin{abstract}
Visual relocalization aims to estimate 6-DoF camera pose from images. It plays an important role in mobile robotics and Augmented/Mixed Reality. 
Absolute Pose Regression (APR) provides an efficient method by directly regressing images to poses without explicit scene databases. However, most existing APR methods primarily depend on appearance-based features and lack explicit geometric information, limiting reliability in texture-sparse or dynamically changing environments. 
DepthLoc is proposed, a depth-inferred visual relocalization framework. DepthLoc leverages a pretrained monocular depth estimator to extract dense geometric information from images. Introduced Geometry-Guided Attention Fusion (GAF) module integrates Depth-based and RGB-based feature, which 
expands the spatial perception capability.
 Building upon the geometry-enriched representations, jointed spatio-temporal graph diffusion is further introduced to get motion-consistent feature representations for pose regression.
Extensive experiments on challenging benchmarks demonstrate that DepthLoc achieves competitive performance across large-scale outdoor self-driving environments. 

\end{abstract}

\section{INTRODUCTION}
Visual relocalization aims to estimate camera pose from the input images. It plays important roles in mobile robotics and augmented/mixed reality. Among existing paradigms of visual relocalization\cite{Brachmann2017DSAC, turkoglu2021relativenet, sarlin2019hloc, keetha2023anyloc}, absolute pose regression(APR)\cite{naseer2017deep, li2024dl, musallam2022leveraging} has gained increasing attention due to its ability to directly infer 6-DoF poses from images without relying on large-scale, explicitly constructed scene databases. APR establishes a compact neural network to directly regress 6-DoF camera poses from images, thereby providing a scalable and real-time alternative to traditional structure-based localization pipelines\cite{sarlin2019hloc,sarlin2021pixloc}.

Over the past decade, APR has emerged as an efficient paradigm for camera relocalization. The initial PoseNet leverages a Convolutional Neural Network(CNN) based on GoogLeNet for pose regression\cite{kendall2015posenet}. Subsequent works such as GeoPoseNet\cite{kendall2017geometric} and LSTM-PoseNet\cite{walch2017poselstm} further demonstrate the feasibility of end-to-end pose prediction by employing geometric loss functions and LSTM modules, respectively. These approaches primarily rely on single-frame appearance information, resulting in weakly constrained geometric and consistency and temporal coherence.

Subsequent works try to combine spatial-temporal aware of multi-frames with APR frameworks. MapNet incorporates relative pose constraints derived from visual odometry to regularize predictions across image sequences, improving temporal consistency and reducing trajectory drift\cite{brahmbhatt2018geomapnet}. Atloc+ further incorporates temporal constraints by introducing attention mechanisms that emphasize globally informative regions, leading to more robust pose estimation in large-scale outdoor environments\cite{wang2020atloc}. Subsequent works\cite{xue2019lsg, xue2020grnet, qiao2023transapr, wang2023robustloc} extend multi-frame APR by performing spatial-temporal modeling of inter-frame relationships through graph-based interactions and diffusion-based interactions, enhancing robustness to viewpoint changes and appearance variations and broadening multi-frame pose regression in long-term robotic applications. 



Although recent advances in APR have improved localization accuracy, most existing approaches handle geometric constraints and temporal consistency through relative pose supervision\cite{qiao2023objects, LI2025IV}. In the previous works, relative pose supervision is applied at the output level rather than embedded within a unified spatiotemporal feature space. As a result, feature processing remains independent, without structurally encoding geometric structure or temporal interaction in feature fusion. This independent structure limits the ability of network to capture the joint interaction of spatial structure and temporal motion. The feature can remain appearance-driven and sensitive to viewpoint changes, illumination variations, and dynamic disturbances\cite{wang2023robustloc}. Geometry and temporal consistency are imposed by loss function rather than emerging from an unified spatio-temporal representation.

To address these limitations, DepthLoc is introduced as a unified geometry-aware spatio-temporal relocalization framework. The introduced Geometry-Guided Attention Fusion (GAF) module enables depth-derived geometric information to explicitly regulate RGB representations through bidirectional cross-attention. Specifically, depth-guided attention emphasizes structurally stable regions, such as building edges and static landmarks, steering the RGB stream toward pose-relevant areas. By conditioning appearance features on geometric structure, sensitivity to transient textures and illumination changes is reduced, and structurally grounded frame-level embeddings that are more discriminative for pose regression are produced.
Building upon these geometry-aware representations, a coupled diffusion mechanism for unified spatio-temporal feature fusion is further introduced. Instead of independently modeling spatial and temporal interactions, multi-modal features are jointly propagated across spatial neighborhoods and temporal sequences within a unified neural diffusion. Through the coupled formulation, consistent feature interaction across frames is encouraged, amplification of short-term appearance noise is suppressed. Consequently, the learned embeddings exhibit temporal consistency for long-term pose estimation.

Experiments on challenging benchmarks demonstrate that DepthLoc achieves competitive performance compared with existing APR methods. By effectively integrating geometric depth priors with appearance features, DepthLoc enhances geometric consistency and temporal stability in pose estimation.

In summary, the contributions of this work are threefold:

\begin{itemize}
\item DepthLoc is proposed, a depth-enhanced APR framework that leverages monocular depth estimation to extract dense geometric priors and introduces a Geometry-Guided Attention Fusion (GAF) module for structured cross-modal interaction, yielding geometrically enriched per-frame representations for robust pose regression.
\item A coupled spatio-temporal diffusion mechanism is introduced that jointly models spatial dependencies and temporal dynamics within a unified formulation, enabling improved motion consistency and more natural pose estimation.
\item Extensive experimental validations show competitive performance across 4Seasons and Oxford Robotcar benchmarks, especially under diverse challenging conditions.
\end{itemize}

\section{Related Work}

\subsection{DNN-based Absolute pose regression}
Absolute pose regression has attracted increasing attention from researchers in recent years. CNNs are widely used to predict camera position and orientation directly from images. The representative model is PoseNet\cite{kendall2015posenet}, which shows that global descriptors of images end-to-end pose estimation. Many subsequent studies refine the idea by improving backbone design and loss formulation. Furthermore, Some methods\cite{kendall2017geometric, moreau2022coordinet} introduce uncertainty modeling or geometry-aware supervision to balance translation and rotation. While these CNN-based approaches are efficient, these methods often depend on local receptive fields, which limits their ability to model large viewpoint changes and repeated structures.

To improve robustness, researchers explore temporal and interaction of multi-view images. multi-frame APR frameworks exploit the relation between adjacent frames to reduce pose error. For example, MapNet\cite{brahmbhatt2018geomapnet} introduces motion consistency across frames by visual odometry, which improves localization accuracy in outdoor large-scale environments. Recurrent neural networks and graph neural network are also used to propagate information over sequences or multi-view graphs\cite{clark2017vidloc, turkoglu2021visual}. Furthermore, Attention mechanisms\cite{wang2020atloc} are introduced to guide the network toward stable regions and suppress dynamic content. These designs enhance spatial consistency and help the model handle moderate scene changes, yet long-range dependencies are still not fully captured.

Recently, transformer architectures have been studied for visual localization due to their global self-attention mechanism\cite{qiao2023transapr, shavit2021MSTrans, xiao2024effloc}. Models such as MS-Trans extend pose regression to transformer-based feature encoding. Lightweight vision transformers are further explored to balance accuracy and efficiency. Global attention\cite{vaswani2017attention} allows direct interaction between distant image regions, which is important for large-scale scenes and complex layouts of environment changes. The architectures with strong global modeling capacity become increasingly relevant for stable and scalable camera pose estimation as the datasets continue to grow in size and diversity.

\subsection{Depth estimation}

Monocular depth estimation infers the distance from each pixel in the scene to the camera from a single RGB image, generating a corresponding dense depth map. Early supervised approaches treat depth prediction as a dense regression task. The convolutional networks learn direct regression from single RGB images to metric depth using large datasets with ground-truth labels from RGB-D sensors or LiDAR. Despite performing well in indoor environments, their accuracy drops considerably when applied to other environments, due to the limited variety of training data.

Monodepth2\cite{godard2019monodepth} leverages full-resolution multi-scale sampling and auto-masking loss from stereo pairs or monocular video as supervision, eliminating the need for ground-truth depth. This paradigm enabled training on unconstrained driving and outdoor sequences, improving robustness to appearance variations. More recently, the field has adopted large foundation model designs. DepthAnythingV2\cite{depth_anything_v2} uses synthetic data for teacher pre-training and large-scale pseudo-labeling to produce fine-grained predictions with strong robustness. Metric3Dv2\cite{hu2024metric3d} resolves metric ambiguity across different camera models by proposing a camera space transformation module, and introduces joint depth-normal optimization to leverage diverse data for improving zero-shot normal estimation. 

\section{Method}

\begin{figure*}[!t]
\centering
\includegraphics[width=0.87\textwidth,height=0.28\textheight]{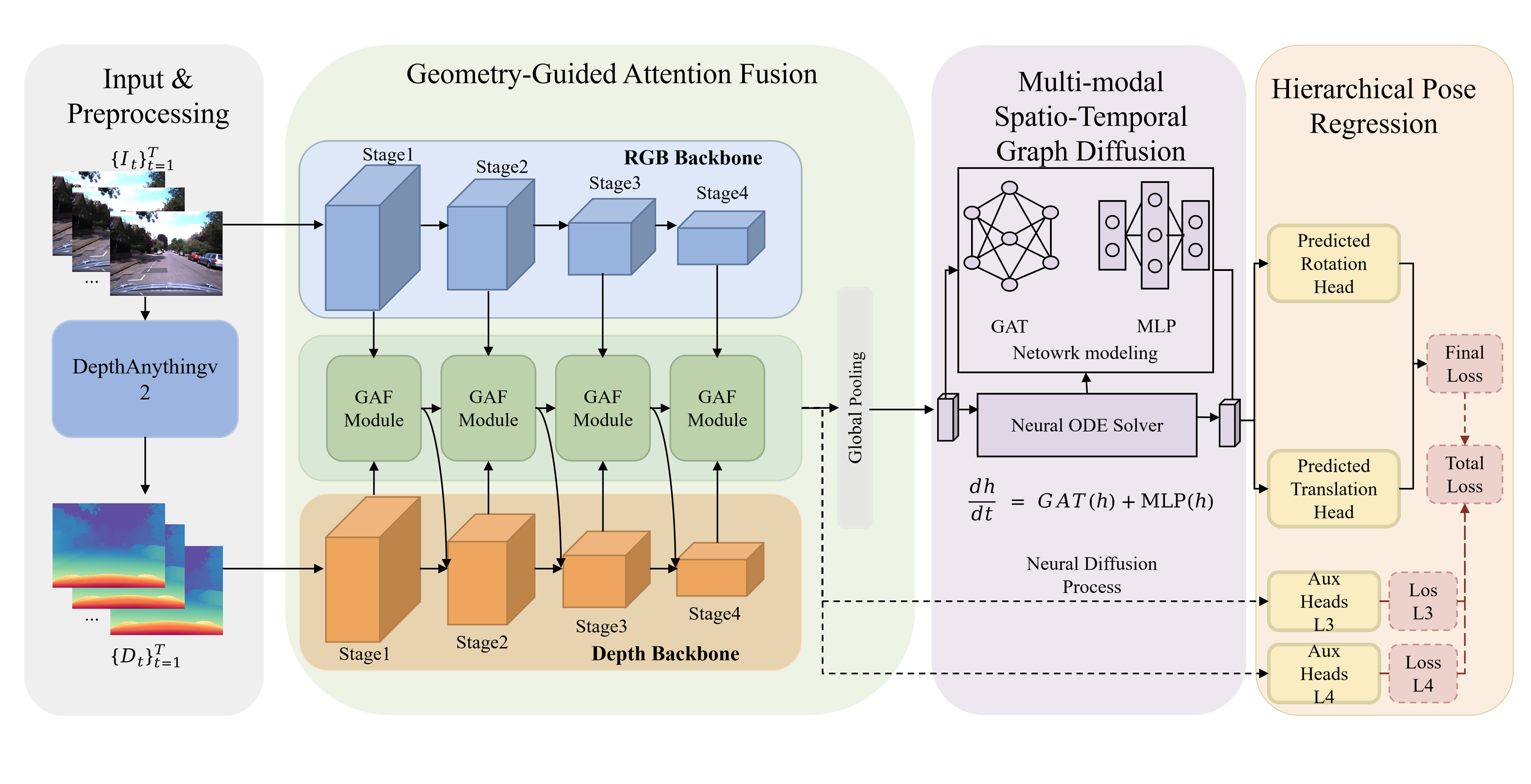}
\centering
\caption{Overview of the proposed DepthLoc framework. The architecture consists of a dual-stream backbone that extracts hierarchical appearance and geometric features. The Geometry-Guided Attention Fusion (GAF) module integrates these features to enhance spatial perception. A spatio-temporal graph attention diffusion mechanism is then employed to model motion consistency across the sequence, followed by hierarchical pose regressor heads that predict the 6-DoF camera poses.}
\label{fig:fig2}
\end{figure*}

This section presents the proposed DepthLoc framework in detail, as illustrated in Fig.\ref{fig:fig2}. 
DepthLoc is designed to unify geometric perception and multi-modal spatio-temporal interaction within an absolute pose regression paradigm. 
Given an RGB image sequence and its corresponding monocular depth maps, the framework first extracts appearance and geometric features through a dual-stream architecture. 
Next, the Geometry-Guided Attention Fusion (GAF) module performs structured cross-modal interaction to generate geometry-appearance representations. 
Subsequently, a spatial-temporal graph diffusion models multi-modal geometry-appearance representations in a unified formulation. 
Finally, hierarchical pose regression heads produce 6-DoF camera pose predictions under multi-level supervision.

\subsection{Dual-Stream RGB and Depth Backbones}

Given an RGB image sequence $\{I_t\}_{t=1}^{T}$ and its corresponding monocular depth maps $\{D_t\}_{t=1}^{T}$, The objective is to regress the 6-DoF camera poses $\{P_t\}_{t=1}^{T}$.

\subsubsection{RGB Backbone}

The RGB branch adopts a ResNet-34 convolutional backbone to extract hierarchical appearance features:

\begin{equation}
\mathbf{F}^{l}_{rgb} = \Phi^{l}_{rgb}(I_t), \quad l \in \{1,2,3,4\},
\end{equation}

where $\Phi^{l}_{rgb}(\cdot)$ denotes the $l$-th residual stage. The branch captures texture, semantic cues, and global scene layout, which are essential for coarse pose inference. 
However, appearance-only representations may suffer from perceptual aliasing and instability in low-texture or dynamic environments.

\subsubsection{DepthAnythingv2 and Depth Backbone}

To introduce explicit geometric priors, a pre-trained DepthAnythingv2 is deployed model to get the geometric information, a parallel backbone with identical architecture processes the depth maps:
\begin{align}
D_t =  \Phi_{d}(I_t), \\
\mathbf{F}^{l}_{d} = \Phi^{l}_{d}(D_t).
\end{align}

Unlike conventional APR methods that rely solely on RGB features, 
the depth stream encodes scene geometry and structural continuity of depth information. 
Such geometric information provides complementary information that improves localization robustness, especially under illumination changes or repetitive textures.

The two streams share aligned hierarchical structures, enabling structured cross-modal interaction at multiple levels.

\subsection{Geometry-Guided Attention Fusion}

\begin{figure}[!t]
\centering
\includegraphics[width=0.48\textwidth,height=0.20\textheight]{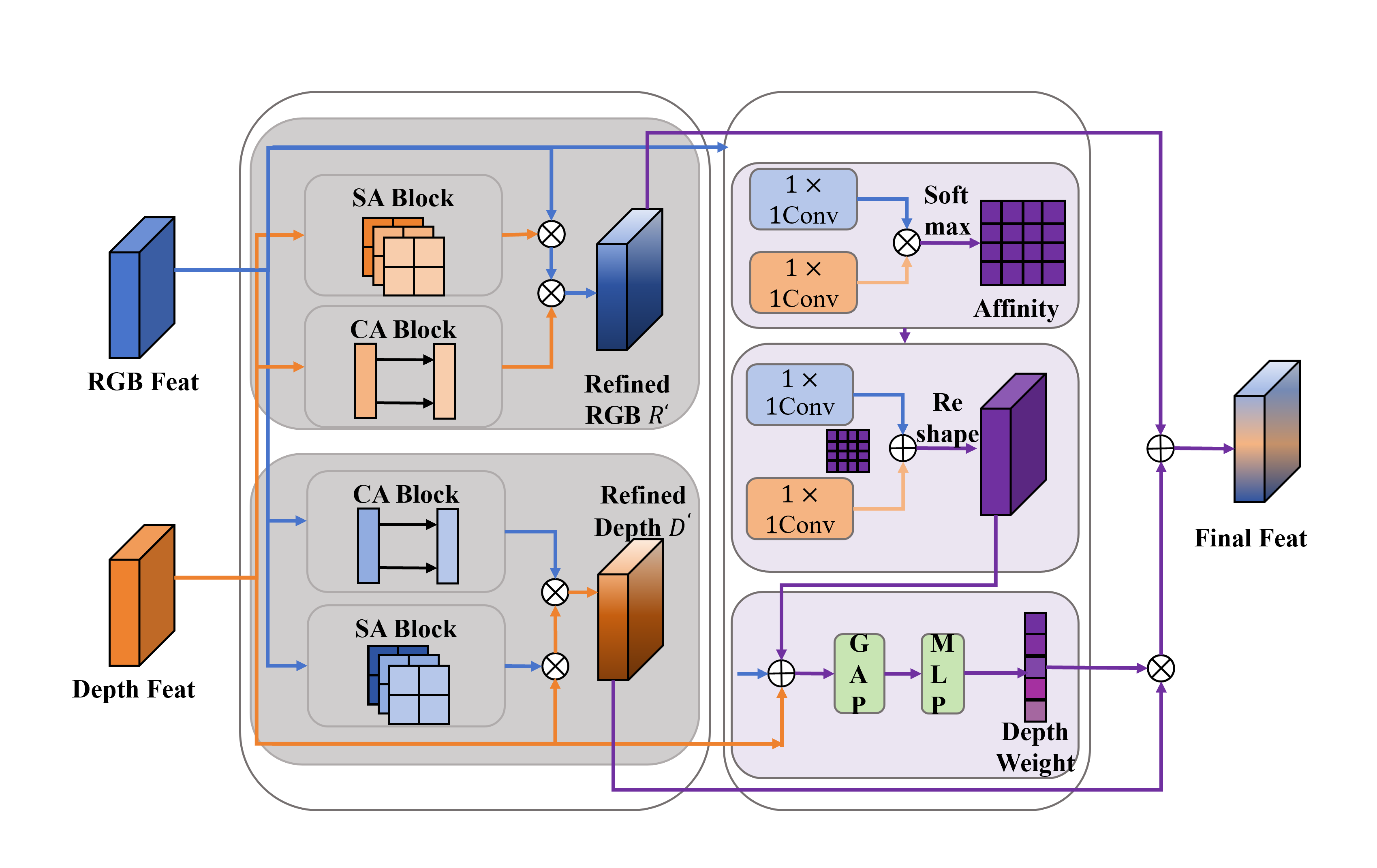}
\centering
\caption{Detailed structure of the Geometry-Guided Attention Fusion (GAF) module. This module performs bidirectional cross-modal interaction, spatial and channel attention from the depth stream modulate the RGB features. An adaptive channel-wise weighting mechanism is then applied to aggregate the enhanced multi-modal features, producing a geometry-consistent representation for each frame.}
\label{fig:fig3}
\end{figure}

While RGB features encode rich semantics, depth features provide geometric constraints. 
Inspired by multi-modal fusion\cite{liu2024depth, wang2025foundation} to effectively integrate these complementary information, a GAF module is introduced that performs structured cross-modal interaction, as illustrated in Fig. \ref{fig:fig3}.

Given RGB features $\mathbf{F}^{l}_{rgb}$ and depth features $\mathbf{F}^{l}_{d}$, fusion is conducted in two stages.
First, spatial and channel attention maps are computed from each modality to modulate the other:

\begin{align}
\mathbf{R}' &= (\text{SA}(\mathbf{F}_d) \odot \mathbf{F}_{rgb}) \odot \text{CA}(\mathbf{F}_d), \\
\mathbf{D}' &= (\text{SA}(\mathbf{F}_{rgb}) \odot \mathbf{F}_d) \odot \text{CA}(\mathbf{F}_{rgb}),
\end{align}
where SA and CA denote spatial and channel attention, and $\odot$ represents element multiplication.
The bidirectional modulation enables depth-aware refinement of RGB representations while allowing appearance information to guide depth feature selection. 
Consequently, salient geometric structures and informative visual information are mutually enhanced.

To prevent depth estimates from adversely affecting pose regression, an adaptive channel-wise weighting mechanism is defined as:
\begin{align}
\mathbf{F}_a &= \text{softmax}(\mathcal{R}(\text{Conv}(\mathbf{F}_d)) 
\mathcal{R}(\text{Conv}(\mathbf{F}_{rgb}))^\top), \\
\mathbf{F}_{a2} &= \mathbf{F}_a \cdot \text{Conv}(\mathbf{F}_d)
+ \mathbf{F}_a \cdot \text{Conv}(\mathbf{F}_{rgb}),
\end{align}
where $\mathcal{R}(\cdot)$ reshapes features into channel form. A global descriptor is then obtained:
\begin{equation}
\mathbf{q}_{id} = \text{MLP}(\text{GAP}(\mathbf{F}_{a2} + \mathbf{F}_d + \mathbf{F}_{rgb})).
\end{equation}

The final fused representation is computed as:
\begin{equation}
\mathbf{F}_{fused} = \mathbf{q}_{id} \odot \mathbf{D}' + \mathbf{R}'.
\end{equation}

The adaptive weighting mechanism ensures geometry-consistent enhancement while maintaining robustness to imperfect depth. 
Overall, the GAF module produces per-frame representations that are both semantically discriminative and geometrically structured.

\subsection{Spatio-Temporal Graph Attention Diffusion}

Single-frame regression lacks motion consistency constraints and may lead to trajectory drift. 
To address this issue, fused multi-modal embeddings is extended into a spatio-temporal diffusion framework for structured information interaction.

The cross-modal fusion allows RGB and depth features to complement each other. 
RGB images may become ambiguous in textureless regions, while depth maps can contain noise or inaccurate estimates. 
Their interaction leverages semantic cues from RGB and geometric structure from depth, yielding more reliable frame-level representations.

After global pooling, the embeddings are represented as:
\begin{equation}
\mathbf{X} = [\mathbf{f}_1, \dots, \mathbf{f}_T].
\end{equation}

Inspired by sequence-based localization strategies, feature interaction as a neural diffusion process\cite{wang2023robustloc, kang2021stable_ode}, governed by a graph attention transformer:
\begin{equation}
\frac{d\mathbf{X}(t)}{dt}
=
\underbrace{\text{GAT}(\mathbf{X}(t))}_{\text{spatial interaction}}
+
\underbrace{\text{MLP}(\mathbf{X}(t))}_{\text{temporal evolution}}.
\end{equation}

The graph attention transformer captures relational dependencies across frames and discovers structural consistency within the sequences, 
while the MLP captures the temporal dynamics of features.
Through this coupled interaction, complementary strengths of both modalities are preserved, 
the temporal logic of the sequence is reinforced, 
and the network learns to maintain the correct ordering of frames while suppressing inconsistent transitions.

The system is solved using neural diffusion:
\begin{equation}
\mathbf{X}(t_1) = 
\int_{t_0}^{t_1}
\left(
\text{GAT}(\mathbf{X}(t)) +
\text{MLP}(\mathbf{X}(t))
\right) dt,
\end{equation}
which promotes smooth feature propagation and reduces trajectory drift.

\subsection{Hierarchical Pose Regression}

To enhance optimization stability and encourage pose-aware feature learning at different semantic depths, 
a hierarchical regression strategy is adopted, which generates pose predictions at multiple feature levels. 
Instead of relying solely on the final sequence embedding, 
the network supervises intermediate fused representations to progressively guide geometry-sensitive encoding.

Specifically, the features of fused Layer 4 are first globally pooled and directly mapped to six pose components through the independent linear heads:
\begin{equation}
\hat{\mathbf{P}}_{l4} = 
\left[
f^{l4}_1(\mathbf{x}_{l4}), \dots, f^{l4}_6(\mathbf{x}_{l4})
\right],
\end{equation}
In parallel, Layer 3 fused features are supervised in the same manner:
\begin{equation}
\hat{\mathbf{P}}_{l3} = 
\left[
f^{l3}_1(\mathbf{x}_{l3}), \dots, f^{l3}_6(\mathbf{x}_{l3})
\right],
\end{equation}

For the final pose regression, the Layer 4 features are further refined by the spatio-temporal graph attention module, producing a temporally aggregated embedding $\mathbf{x}$. 
This representation encodes both geometry-aware spatial cues and sequence-level motion dependencies. 
The embedding is decomposed into translation and rotation branches:
\begin{align}
\mathbf{z}_t &= \text{ReLU}(W_t \mathbf{X}), \\
\mathbf{z}_q &= \text{ReLU}(W_q \mathbf{X}),
\end{align}
followed by independent linear predictors:
\begin{equation}
\hat{\mathbf{t}} = [f_1(\mathbf{z}_t), f_2(\mathbf{z}_t), f_3(\mathbf{z}_t)],
\end{equation}
\begin{equation}
\hat{\mathbf{q}} = [g_1(\mathbf{z}_q), g_2(\mathbf{z}_q), g_3(\mathbf{z}_q)].
\end{equation}

The final pose estimate
\begin{equation}
\hat{\mathbf{P}} = [\hat{\mathbf{t}}, \hat{\mathbf{q}}],
\end{equation}
the model aligns shallow, mid-level, and temporal representations with pose estimation objectives, 
thereby improving convergence behavior and regression robustness.
\begin{table*}[!htp]
\caption{Mean and median translation/rotation estimation errors(m/°) on 4seasons dataset. The best results in each metric are highlighted with bold respectively.}
\label{tab2}
\renewcommand{\arraystretch}{0.6}
\setlength{\extrarowheight}{2pt}   
\small  
\resizebox{1\linewidth}{!}{
\begin{tabular}{ccccccc}
\toprule
\textbf{}& \multicolumn{2}{c}{\textbf{Business Campus}} & \multicolumn{2}{c}{\textbf{Neighborhood}} & \multicolumn{2}{c}{\textbf{Old Town}} \\
\textbf{Model}& \textbf{Mean} & \textbf{Median} & \textbf{Mean} & \textbf{Median} & \textbf{Mean} & \textbf{Median}\\
\midrule
GeoPoseNet\cite{kendall2017geometric} & 11.04 / 5.78 & 5.93 / 2.03 & 2.87 / 1.30 & 1.92 / 0.88 & 64.81 / 6.67 & 15.03 / 1.57 \\
MapNet\cite{brahmbhatt2018geomapnet} & 10.35 / 3.78 & 5.66 / 1.83 & 2.81 / 1.05 & 1.89 / 0.92 & 46.56 / 7.14 & 16.52 / 2.12 \\

AtLoc\cite{wang2020atloc} & 10.36 / 5.35 & 5.95 / 2.74 & 4.52 / 2.47 & 3.74 / 1.71 & 64.27 / 8.23 & 23.99 / 2.62 \\
AtLoc+\cite{wang2020atloc} & 13.70 / 6.41 & 5.58 / 1.94 & 2.33 / 1.39 & 1.61 / 0.88 & 68.40 / 5.51 & 14.52 / 1.69 \\
RobustLoc\cite{wang2023robustloc} & 4.28 / \textbf{2.04} & 2.55 / 1.50 & 1.36 / \textbf{0.83} & 1.00 / 0.65 & 21.65 / 2.41 & 5.52 / \textbf{1.05} \\
IRPNet\cite{shavit2021IRPNet} & 10.95 / 5.38 & 5.91 / 1.82 & 3.17 / 2.85 & 1.98 / 0.90 & 55.86 / 6.97 & 17.33 / 3.11 \\
CoordiNet\cite{moreau2022coordinet} & 11.52 / 3.44 & 6.44 / 1.38 & 1.72 / 0.86 & 1.37 / 0.69 & 43.68 / 3.58 & 11.83 / 1.36 \\
GRNet\cite{xue2020grnet} & 7.69 / 4.34 & 5.52 / 2.16 & 3.02 / 2.92 & 2.14 / 1.45 & 41.54 / 7.30 & 19.23 / 3.26 \\
Ours & \textbf{3.73} / 2.32 &  \textbf{2.12} /  \textbf{1.26} &  \textbf{1.16} / 0.86 &  \textbf{0.89} / \textbf{0.64} & \textbf{7.88} /  \textbf{1.71} &  \textbf{4.31} /  1.06 \\
\bottomrule
\end{tabular}
}
\end{table*}

\subsection{Loss Function}

The loss function is designed to balance translation and rotation in a principled way. the model learns the relative scale during training.

Given a predicted pose $\hat{\mathbf{P}} = [\hat{\mathbf{t}}, \hat{\mathbf{q}}]$ 
and ground truth $\mathbf{P} = [\mathbf{t}, \mathbf{q}]$, 
the absolute pose loss is defined as:
\begin{equation}
\mathcal{L}_{abs} =
e^{-s_t} \|\hat{\mathbf{t}} - \mathbf{t}\|_1 + s_t
+
e^{-s_q} \|\hat{\mathbf{q}} - \mathbf{q}\|_1 + s_q.
\end{equation}

Here, $s_t$ and $s_q$ are learnable parameters. 
They control the relative importance of translation and rotation. 
This formulation follows the uncertainty-based weighting strategy. 
It helps stabilize training and avoids manual scale tuning.

add a relative pose constraint is added to enforces motion consistency between adjacent frames. 
This function compute relative motion from consecutive predictions. 
The predicted increments is denoted as $\Delta \hat{\mathbf{P}}$ 
and the ground truth increments as $\Delta \mathbf{P}$.

The relative pose loss is:
\begin{equation}
\mathcal{L}_{vo} =
e^{-s_r} \|\Delta \hat{\mathbf{t}} - \Delta \mathbf{t}\|_1 + s_r
+
e^{-s_q} \|\Delta \hat{\mathbf{q}} - \Delta \mathbf{q}\|_1 + s_q.
\end{equation}

where $s_r$ is also learnable parameter. This term reduces sudden frame-to-frame changes. It improves temporal smoothness. both loss are combined as:
\begin{equation}
\mathcal{L}_{pose} = \mathcal{L}_{abs} + \mathcal{L}_{vo}.
\end{equation}

The network produces predictions at multiple levels. Therefore, Three losses supervise all levels jointly. The final loss is:
\begin{equation}
\mathcal{L}_{total} =
\mathcal{L}_{final}
+
\lambda_3 \mathcal{L}_{l3}
+
\lambda_4 \mathcal{L}_{l4}.
\end{equation}

Here, $\mathcal{L}_{final}$ supervises the sequence-refined prediction. 
$\mathcal{L}_{l3}$ and $\mathcal{L}_{l4}$ supervise the intermediate outputs 
$\mathcal{P}_{l3}$ and $\mathcal{P}_{l4}$. 

This design enforces absolute accuracy and motion consistency at the same time. 
It also improves gradient flow across the hierarchy. 
As a result, the model achieves more stable pose regression.

\section{Experiments}

\subsection{Implementation Details}

DepthLoc is implemented in PyTorch and trained on an NVIDIA
3090 GPU. ResNet34\cite{he2016resnet} backbones pre-trained on ImageNet are adopted for
both the RGB and depth streams. The model takes monocular RGB image sequences as input, while
DepthAnythingv2 (Large) provides the inferred depth maps. The sequence length is set to 7, and each
image is resized to 128 pixels on the shorter side. The batch size is 64.
Optimization uses Adam with a learning rate of $2 \times 10^{-4}$ and
weight decay of $5 \times 10^{-4}$ for 400 epochs. Data augmentation includes color jitter and random
cropping. The diffusion integration steps are set to 0 and 1, and the number of attention heads is 8.

\begin{figure}[!t]
\centering
\includegraphics[width=0.40\textwidth,height=0.32\textheight]{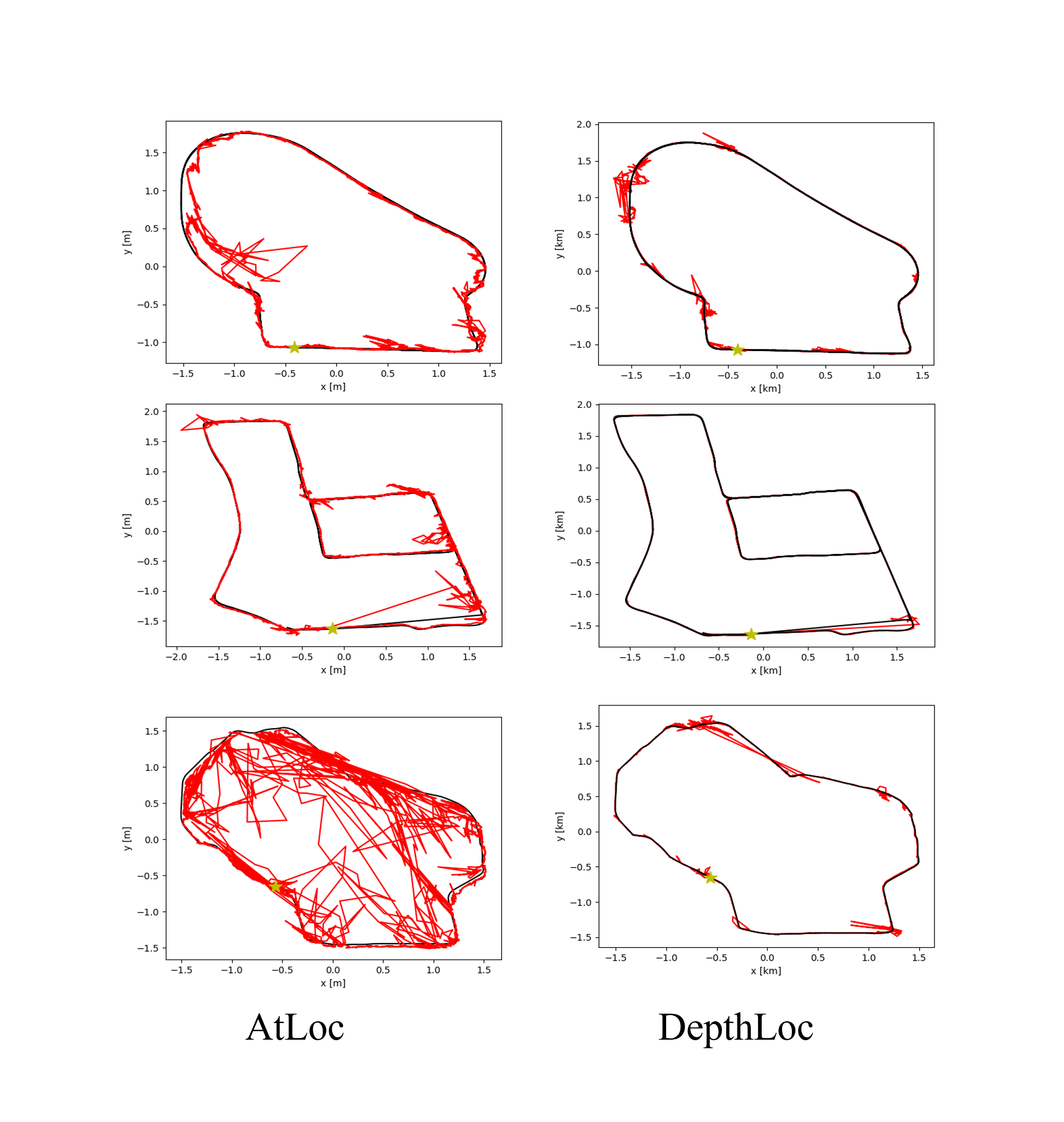}
\centering
\caption{Trajectory predictions on Business Campus(top), Neighborhood(middle), and Old Town(bottom) of 4Seasons Dataset. Ground truth paths are shown in black, predicted trajectories in red, and yellow stars indicate the start points.}
\label{fig:fig4}
\end{figure}

\begin{table*}[!t]
\caption{Mean and median translation/rotation estimation errors(m/°) on Robotcar dataset. The best results in each metric are highlighted with bold respectively.}
\label{tab2}
\renewcommand{\arraystretch}{0.6}
\setlength{\extrarowheight}{2pt}   
\small  
\resizebox{1\linewidth}{!}{
\begin{tabular}{ccccccc}
\toprule
\textbf{}& \multicolumn{2}{c}{\textbf{Loop1}} & \multicolumn{2}{c}{\textbf{Loop2}} & \multicolumn{2}{c}{\textbf{Full1}} \\
\textbf{Model}& \textbf{Mean} & \textbf{Median} & \textbf{Mean} & \textbf{Median} & \textbf{Mean} & \textbf{Median}\\
\midrule
PoseNet+\cite{kendall2017geometric} & 25.39 / 17.49 & 6.91 / 2.06 & 28.89 / 19.65 & 5.83 / 2.05 & 125.6 / 28.10 & 107.8 / 23.5 \\
MapNet\cite{brahmbhatt2018geomapnet} & 8.79 / 3.53 & 5.79 / 1.54 & 9.81 / 3.86 & 4.93 / 1.67 & 41.2 / 13.5 & 17.91 / 6.39 \\

AtLoc\cite{wang2020atloc} & 8.61 / 4.58 & 5.68 / 2.23 & 8.86 / 4.67 & 5.05 / 2.01 & 29.6 / 12.4 & 11.1 / 5.28 \\
AtLoc+\cite{wang2020atloc} & 7.82 / 3.62 & - / - & 7.24 / 3.60 & - / - & 21.0 / 6.15 & - / - \\
LsG\cite{xue2020grnet} & 9.07 / 3.31 & - / - & 9.19 / 3.53 & - / - & 31.65 / 4.51 & - / - \\
GRNet\cite{xue2020grnet} & 7.76 / \textbf{2.54} & - / - & 8.15 / \textbf{2.57} & - / - & 17.35 / 3.47 & - / - \\
Effloc\cite{wang2023robustloc} & 7.58 / 3.72 & 5.23 / 2.18 & 7.89 / 4.19 & 4.76 / 2.06 & 27.23 / 11.41 & 11.12 / 4.17 \\
Ours & \textbf{5.31} / 2.80 &  \textbf{4.94} /  \textbf{1.49} &  \textbf{5.15} / 2.89 &  \textbf{4.43} / \textbf{1.53} & \textbf{13.05} / \textbf{4.11} &  \textbf{7.10} / \textbf{1.50} \\
\bottomrule
\end{tabular}
}
\end{table*}

\begin{figure}[!t]
\centering
\includegraphics[width=0.50\textwidth,height=0.28\textheight]{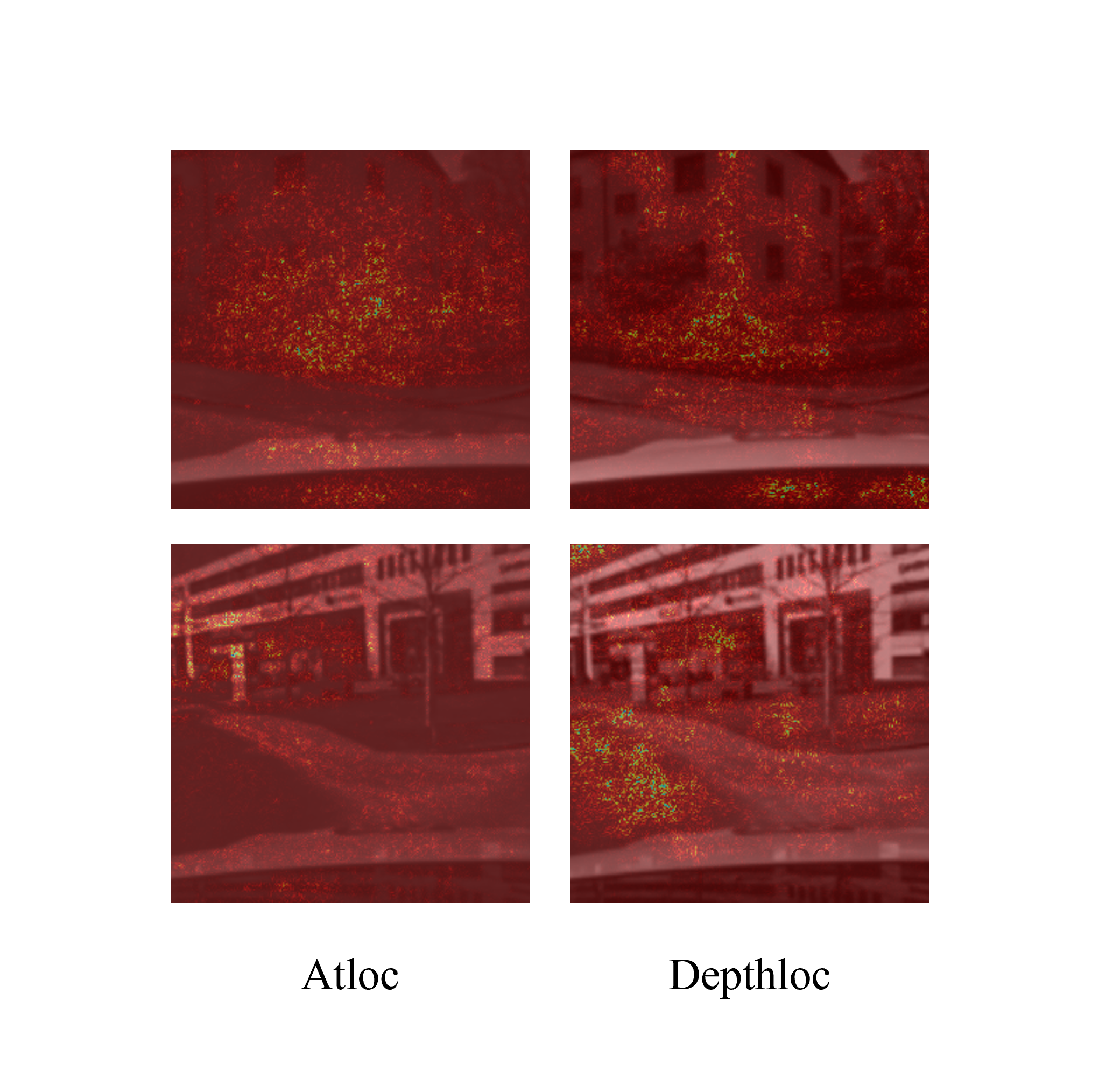}
\centering
\caption{Saliency maps on Business Campus(top) and Neighborhood(bottom) of Oxford 4Seasons Dataset.}
\label{fig:fig5}
\end{figure}

\subsection{Datasets}
\textbf{4Seasons\cite{wenzel20214seasons}:} The 4Seasons dataset is a challenging outdoor benchmark captured across different seasons and weather conditions. It contains multiple sequences with significant appearance variations due to changing lighting and weather changes. DepthLoc is evaluated on three representative scenes: \textit{business\_campus}, \textit{neighborhood}, and \textit{old\_town}, each scene presents unique challenges such as repetitive patterns, dynamic objects, and narrow streets with illumination changes.

\textbf{Oxford RobotCar\cite{maddern20171_robotcar}:} The Oxford RobotCar dataset contains image sequences acquired over 100 repetitions of a continuous route through central Oxford over a year. the evaluation  performance of DepthLoc on both LOOP and FULL scenarios to assess its effectiveness and robustness under varying scales and environmental conditions.

\begin{table}[htbp]
    \centering
    \caption{Dataset configurations for Oxford RobotCar and 4Seasons.}
    \label{tab:dataset_config}
    \resizebox{\linewidth}{!}{%
    \begin{tabular}{llll}
        \toprule
        \multirow{2}{*}{Dataset} & \multirow{2}{*}{Sequence} & \multicolumn{2}{c}{Data Split} \\
        \cmidrule(lr){3-4}
        & & Train & Test \\
        \midrule
        \multirow{7}{*}{\textbf{Oxford RobotCar}} & Loop1 & 2014-06-26-08-53-56 & 2014-06-23-15-36-04 \\
        & Loop2 & 2014-06-26-09-24-58 & 2014-06-23-15-41-25 \\
        \cmidrule(lr){2-4}
        & \multirow{3}{*}{Full} & 2014-11-28-12-07-13 & 2014-12-09-13-21-02 \\
        & & 2014-12-02-15-30-08 & \\

        \midrule
        \multirow{10}{*}{\textbf{4Seasons}} & \multirow{3}{*}{Business Campus} & 2020-10-08--09-30-57 & 2021-02-25--14-16-43 \\
        & & 2021-01-07--13-12-23 & \\
        \cmidrule(lr){2-4}
        & \multirow{4}{*}{Neighborhood} & 2020-03-26--13-32-55 & 2021-05-10--18-32-32 \\
        & & 2020-10-07--14-47-51 & \\
        & & 2020-10-07--14-53-52 & \\
        & & 2020-12-22--11-54-24 & \\
        \cmidrule(lr){2-4}
        & \multirow{3}{*}{Old Town} & 2020-10-08--11-53-41 & 2021-02-25--12-34-08 \\
        & & 2021-01-07--10-49-45 & \\
        & & 2021-05-10--21-32-00 & \\
        \bottomrule
    \end{tabular}
    }
\end{table}

\begin{figure*}[!t]
\centering
\includegraphics[width=0.82\textwidth,height=0.31\textheight]{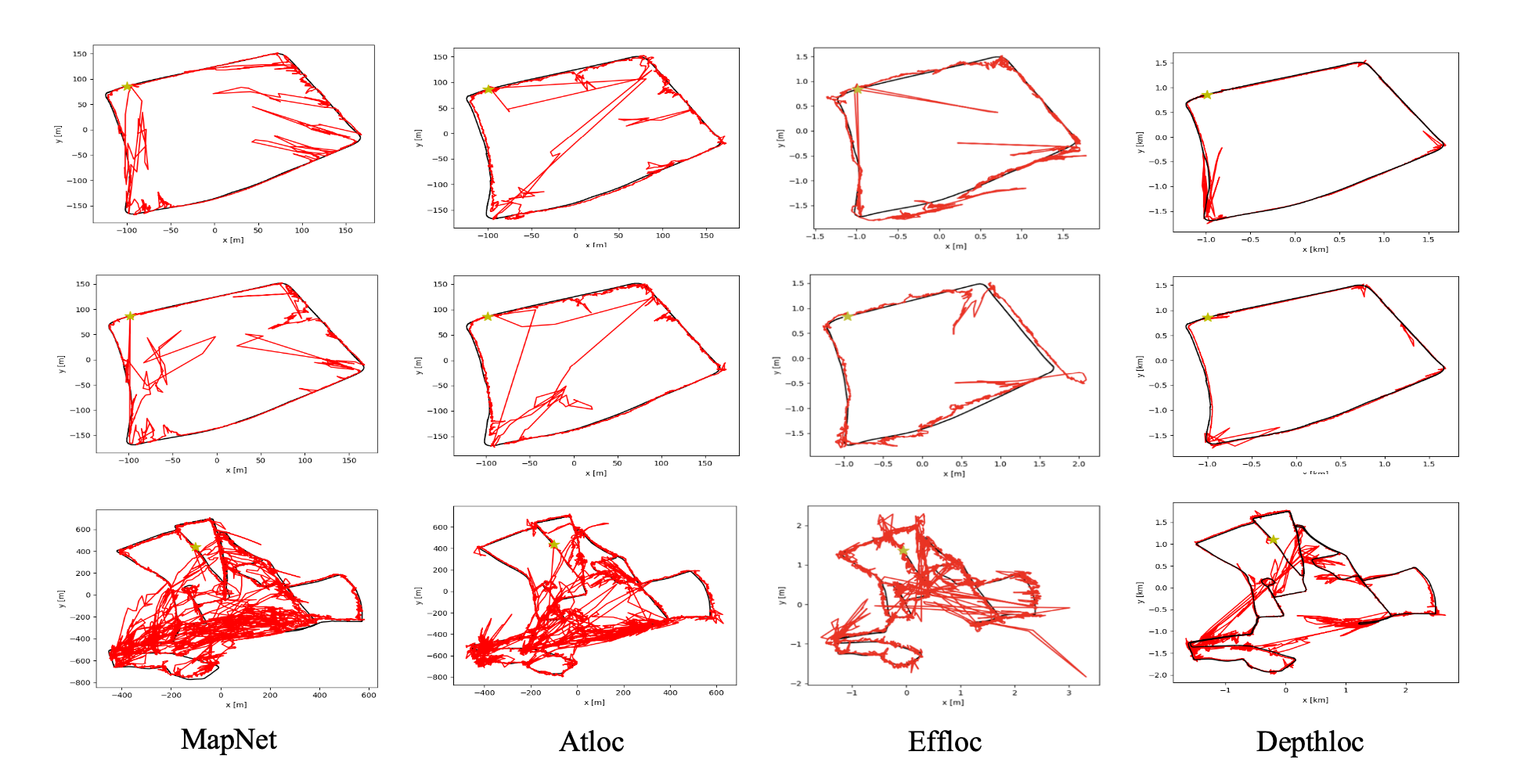}
\centering
\caption{Trajectory predictions on Loop1(top), Loop2(middle), and Full1(bottom) of Robotcar Dataset. Ground truth paths are shown in black, predicted trajectories in red, and yellow stars indicate the start points.}
\label{fig:fig6}
\end{figure*}

\subsection{Experiments on the 4Seasons Dataset}

The 4Seasons dataset presents considerable challenges due to drastic seasonal and illumination changes, necessitating a relocalization model with strong robustness to appearance variations. Table I reports the quantitative comparison of DepthLoc against several APR methods on three representative scenes. DepthLoc achieves competitive performance across nearly all metrics, indicating the benefits of integrating geometric depth cues with spatio‑temporal modeling.

In the Business Campus scene, DepthLoc obtains a mean translation error of 3.73m and a median translation of 2.12m, the median rotation error is reduced from 1.50° to 1.26°. DepthLoc achieves better localization accuracy, particularly on curved roads and at turns. In such relatively open areas where texture information is not highly distinct, these specific scenarios tend to provide significant depth and appearance cues for reliable pose estimation. In the Neighborhood scene, DepthLoc obtains a mean translation error of 1.16m and a median of 0.89m, compared to RobustLoc's 1.36m and 1.00m. In the Neighborhood scene, vehicle typically moves between houses that offer a greater variety of appearance information. As a result, the improvement brought by geometric information is relatively limited compared to Robustloc. In Old Town scene, characterized by narrow streets and illumination change, DepthLoc reduces the mean translation error from 21.65m to 7.88m, and the median translation error from 5.52m to 4.31m, vehicle generally moves through the main city streets, which include both open squares and narrow lanes. With drastic illumination changes and numerous dynamic objects, the integration of depth information plays a crucial role in ensuring localization stability under such challenging conditions.

These improvements can be attributed to two key components of DepthLoc. First, the Geometry‑Guided Attention Fusion module incorporates dense geometric priors from monocular depth estimation. Second, the spatio‑temporal graph attention diffusion encourages motion consistency across frames, contributing to the suppression of outliers caused by dynamic objects and illumination changes, as suggested by the trajectories shown in Fig.\ref{fig:fig4}.

As shown in Fig. \ref{fig:fig5}, in some turning scenarios of the 4Seasons dataset, DepthLoc's attention mechanism, guided by depth cues, focuses on more salient structural regions compared to AtLoc. In the Business Campus scene, DepthLoc's attention focuses more on the overall structure of the buildings, whereas AtLoc's attention, although also on the buildings, is irrelevant to the building structures themselves. In the Neighborhood scene, AtLoc's attention is relatively small and manifests on the edges of vanishing buildings, while DepthLoc focuses on the edges between the buildings in the upper-left corner and the sky, with more distribution on the roads and buildings.

\subsection{Experiments on the Oxford RobotCar Dataset}
Table II compares DepthLoc with several baselines on the Loop1, Loop2, and the Full1 sequences of the Oxford RobotCar dataset.
In Loop1, DepthLoc reduces the mean translation error from 7.58m  to 5.31m, and the median rotation error is lowered from 1.54° to 1.49°. In Loop2, the mean translation error is reduced from 7.89m to 5.15m, and the median rotation error falls from 1.67° to 1.53°. In the Loop scene, DepthLoc achieves competitive performance in both translation and rotation, suggesting stable geometric modeling under illumination changes. In comparison, several prior methods show higher mean errors, likely due to increased sensitivity to illumination changes.

In the Full scene, DepthLoc reduces the mean translation error from 17.35m to 13.05m. The median translation error decreases from 11.12m to 7.10m. The mean rotation error is lowered from 11.41° to 4.11°, and the median rotation error drops from 4.17° to 1.50°. In the more challenging Full setting, DepthLoc obtains lower mean translation error than most baselines, indicating improved robustness under stronger appearance changes and dynamic objects. Nevertheless, performance degradation is still observed in long straight segments and far-range dominant areas, where limited geometry and appearance may lead to translation drift. DepthLoc achieves comparable results to GRNet on the Full scenario. GRNet employs GNN for information interaction, a process similar to the diffusion mechanism in DepthLoc. However, it does not incorporate depth fusion module, leading to localization outliers in textureless and Lighting-variant scenes.

The trajectory visualizations in Fig.\ref{fig:fig6}. Loop scenes further indicate that DepthLoc recovers smooth trajectories with reduced drift, while PoseNet produces numerous outliers and MapNet and GRNet exhibit noticeable drift.

\subsection{Ablation Study}

To assess the contribution of each component in DepthLoc, ablation studies are conducted on the Loop Scene of RobotCar dataset, with results summarized in Table IV. mean translation and rotation errors are reported on the Loop scenes and average the values across Loop scenes. Configuration (a) uses only RGB features without depth fusion or temporal modeling, feeding CNN features directly into the pose regressors. Configurations (b) and (c) add Depth-wise element addition and the GAF module, respectively. Both additions reduce translation and rotation errors, indicating their individual effectiveness. The combined setting in (d) yields the best overall performance, confirming the complementary benefits of cross-modal fusion and temporal graph attention diffusion.

\begin{table}[t]
\centering
\caption{Ablation study of Depth, GAF, and Diffusion modules on Loop scene of Robotcar Dataset. Lower is better.}
\label{tab:ablation}
\begin{tabular}{c|c c c|c c}
\hline
ID & Depth Augmentation & GAF & Diffusion & $t$ (m) $\downarrow$ & $r$ ($^\circ$) $\downarrow$ \\
\hline
A & $\times$ & $\times$ & $\times$ & 10.64 & 3.84 \\ 
B & $\checkmark$ & $\times$ & $\times$ & 7.39 & 3.27 \\ 
C & $\checkmark$ & $\checkmark$ & $\times$ & 6.53 & 3.14 \\
D & $\checkmark$ & $\checkmark$ & $\checkmark$ & \textbf{5.23} & \textbf{2.85} \\ 
\hline
\end{tabular}
\end{table}

\section{CONCLUSIONS}

This paper presents DepthLoc, a depth-enhanced visual relocalization framework that integrates geometric information into absolute pose regression via monocular depth estimation. The Geometry-Guided Attention Fusion module enables effective cross-modal interaction between RGB and depth features, mitigating perceptual aliasing in texture-sparse environments. The spatio-temporal graph attention diffusion mechanism enforces motion consistency across sequences, reducing trajectory drift and improving robustness to dynamic objects and illumination changes. Experiments on 4Seasons and Oxford RobotCar demonstrate that DepthLoc achieves competitive performance against existing APR methods, with notable improvements on challenging sequences. Ablation studies validate the complementary contributions of geometric fusion and temporal modeling. Future work may explore online depth refinement for unseen environments.





\bibliographystyle{unsrt}
\bibliography{ref.bib}

\end{document}